\documentclass[pdflatex,sn-mathphys-num]{sn-jnl}

\usepackage{graphicx}%
\usepackage{multirow}%
\usepackage{amsmath,amssymb,amsfonts}%
\usepackage{amsthm}%
\usepackage{mathrsfs}%
\usepackage[title]{appendix}%
\usepackage{xcolor}%
\usepackage{textcomp}%
\usepackage{manyfoot}%
\usepackage{booktabs}%
\usepackage{algorithm}%
\usepackage{algorithmicx}%
\usepackage{algpseudocode}%
\usepackage{listings}%

\theoremstyle{thmstyleone}%
\theoremstyle{thmstyletwo}%

\theoremstyle{thmstylethree}%

\newcommand{\sth}[1]{\textcolor{black}{#1}}

\begin{document}

\title[Article Title]{Scale over Preference: The Impact of AI-Generated Content on Online Content Ecology}


\author[1]{\fnm{Tianhao} \sur{Shi}}
\email{sth@mail.ustc.edu.cn}
\equalcont{These authors contributed equally to this work.}

\author[2]{\fnm{Yang} \sur{Zhang}}
\email{zyang1580@gmail.com}
\equalcont{These authors contributed equally to this work.}

\author[2]{\fnm{Xiaoyan} \sur{Zhao}}
\email{xzhao@se.cuhk.edu.hk}

\author[2]{\fnm{Fengbin} \sur{Zhu}}
\email{fengbin@nus.edu.sg}

\author[3]{\fnm{Chenyi} \sur{Lei}}
\email{leichenyi@gmail.com}

\author[3]{\fnm{Han} \sur{Li}}
\email{lihan08@kuaishou.com}

\author[3]{\fnm{Wenwu} \sur{Ou}}
\email{ouwenwu@gmail.com}

\author[4]{\fnm{Tian} \sur{Yang}}
\email{tyang@cuhk.edu.hk}

\author[3]{\fnm{Yang} \sur{Song}}
\email{ys@sonyis.me}

\author*[1]{\fnm{Yongdong} \sur{Zhang}}
\email{zhyd73@ustc.edu.cn}

\author*[1]{\fnm{Fuli} \sur{Feng}}
\email{fengfl@ustc.edu.cn}

\affil[1]{\orgname{University of Science and Technology of China}, \city{Hefei}, \country{China}}
\affil[2]{\orgname{National University of Singapore}, \country{Singapore}}
\affil[3]{\orgname{Kuaishou Technology}, \city{Beijing}, \country{China}}
\affil[4]{\orgname{The Chinese University of Hong Kong}, \city{Hong Kong}, \country{China}}

\abstract{
The rapid proliferation of Artificial Intelligence-Generated Content (AIGC) is fundamentally restructuring online content ecologies, necessitating a rigorous examination of its behavioral and distributional implications. Leveraging a comprehensive longitudinal dataset comprising tens of millions of users from a leading Chinese video-sharing platform, this study elucidated the distinct creation and consumption behaviors characterizing AIGC versus Human-Generated Content (HGC). We identified a prevalent \textit{scale-over-preference} dynamic, wherein AIGC creators achieve aggregate engagement comparable to HGC creators through high-volume production, despite a marked consumer preference for HGC. Deeper analysis uncovered the ability of the algorithmic content distribution mechanism in moderating these competing interests regarding AIGC. These findings advocated for the implementation of AIGC-sensitive distribution algorithms and precise governance frameworks to ensure the long-term health of the online content platforms.
}

\keywords{AIGC, Online content ecosystems, Algorithmic recommendation, User engagement dynamics, Distributional effects, Platform governance}



\maketitle

\section{Introduction}\label{sec1}

\begin{figure}[H]
    \centering
    \includegraphics[width=0.975\linewidth]{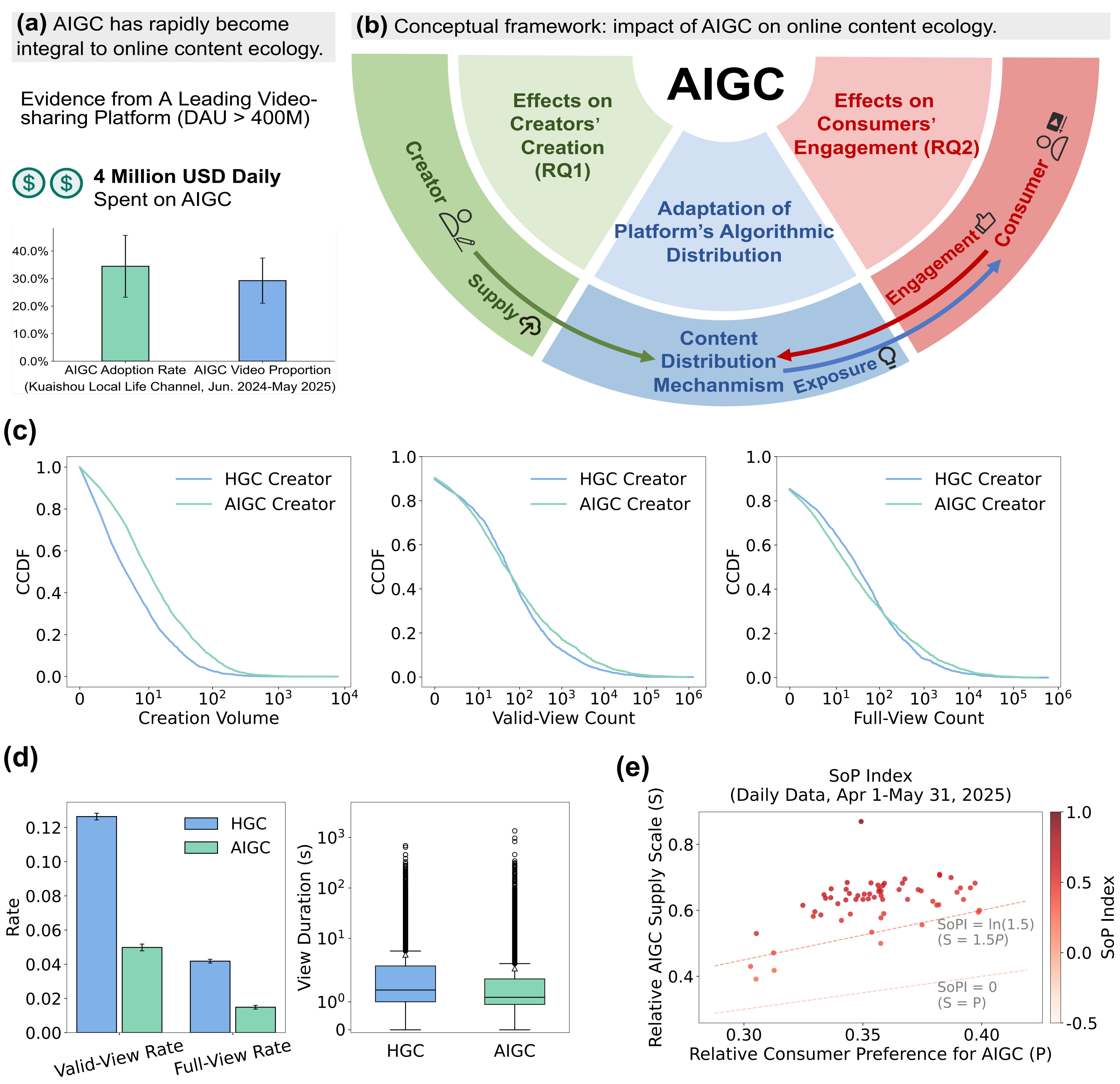}
    \caption{
    \textbf{The scale-over-preference dynamic of AIGC in content creation and consumption.}
\sth{\textbf{(a) Integration of AIGC into the studied platform.} The studied platform has invested heavily in generative AI, with daily expenditures on AIGC reaching 4 million USD across the entire platform \cite{Kuaishou2025Q1}. Specifically within its Local Life channel, about 35\% of creators have adopted AIGC tools, and AIGC videos account for ~30\% of total uploads on average.}
\textbf{(b) Conceptual framework.} Schematic illustration of how AIGC reshapes the content ecology through changes in creator production, consumer engagement, and algorithmic distribution.
\textbf{(c) Creator production and engagement returns.} Complementary cumulative distribution function (CCDF) plots for matched creators ($n = 2,497$ pairs) show that AIGC creators exhibit substantially higher productivity (rightward shift in video volume; left panel) while achieving aggregate engagement returns—measured by total valid views (middle) and full views (right)—comparable to those of HGC creators. Valid views denote views exceeding the platform duration threshold, and full views denote videos watched to completion.
\textbf{(d) Consumer engagement preference.} Matched interaction pairs ($n = 47,288$ \sth{pairs}) reveal systematically lower engagement with AIGC than HGC. AIGC interactions yield lower mean valid-view and full-view rates (left) and shorter view duration (right). Left subpanels report means with 95\% CIs; right boxplots show median (line), mean (triangle), IQR (box), and whiskers at $1.5\times$ IQR.
 \textbf{(e) Scale-over-Preference index (SoPI) for each day.} Daily observations ($n = 61$ days) map relative AIGC supply scale ($S$) against relative consumer preference on AIGC ($P$), colored by SoP index 
$SoPI
=ln(S/P)$. Most days fall in the high-supply/low-preference region above the 
$SoPI=ln(1.5)$ contour (dashed), indicating a persistent divergence between production scale and observed preference.
}
    \label{fig:results_1}
\end{figure}

%
%
%
%
%
%
AIGC is restructuring online content ecologies~\cite{aigc-effect,labeling,moller2026impact}, propelled by the success of generative models~\cite{instructGPT,Wang2026-xq,deepseek,esser2024scaling,hong2023cogvideo}. Accordingly, leading content platforms have elevated AIGC as a strategic priority~\cite{MetaQ42024,Kuaishou2025Q1}. 
Notably, a leading Chinese video-sharing platform, with over 400 million daily active users, has officially launched tools to facilitate AIGC creation. By mid-2025, AIGC has accounted for nearly 30\% of its Local Life Channel (Fig. \ref{fig:results_1}(a)).
Despite this proliferation, the implications of AIGC on human behavior and content distribution remain insufficiently explored. 
Given that these factors determine the flow of attention and rewards~\cite{dai2023effect,petersen2025university}, understanding their dynamics is essential for the development of effective distribution algorithms and governance frameworks.


%
%
%
To bridge this gap, we conducted a large-scale retrospective analysis encompassing hundreds of millions of videos from this platform, spanning 12 months with substantial AIGC volume. 
Framed by the understanding that platforms connect creators and consumers with interest-based content distribution algorithms (Fig.~\ref{fig:results_1}(b))~\cite{qian2024digital,ncf,chen2024macro,zhang2019deep}, we investigated the implications of AIGC from both the behavioral and distributional perspectives through matching~\cite{matching} and temporal analysis~\cite{stock1993simple}. 
We first characterized the systematic disparities in creation and consumption patterns between AIGC and HGC (RQ1 and RQ2), and subsequently examined how algorithmic distribution mechanisms adapted to these AIGC-driven behavioral shifts (RQ3).

%
%
%
Our analyses revealed a \textit{scale-over-preference} dynamic wherein AIGC creators produce higher volumes of content that achieve overall engagement returns comparable to HGC creators, despite weaker individual consumer preferences for AIGC. This divergence signals competing incentives: creators may maximize engagement returns by proliferating AIGC, thereby diluting the density of content that aligns with consumer preferences. Furthermore, we uncovered that algorithmic distribution mechanisms moderate this tension by dynamically adjusting exposure distributions, preserving balanced engagement for creators as AIGC scales. These results provide a mechanistic understanding of how AIGC reshapes online content ecologies, offering insights for sustaining ecosystem health and guiding platform governance in the generative AI era.

\section{Results}

We analysed data from a leading Chinese video-sharing platform with over 400 million daily active users. In early 2024, the platform deployed its proprietary AIGC tool in its Local Life channel. From June 2024 to May 2025, approximately 35\% of active creators adopted the tool, and AIGC videos constituted about 30\% of total content in the channel (Fig.~\ref{fig:results_1}(a)). This large-scale adoption produced a marked shift in content supply and consumption, providing a real-world context to evaluate AIGC's ecological consequences on creator behaviour, consumer engagement, and algorithmic distribution.

\subsection{Scale-over-Preference Dynamics of AIGC}
\label{sec:sop}

We examined the behavioral consequences of AIGC by comparing content creation and consumption patterns between AIGC and HGC. Using two months of data from the platform's Local Life channel (April–May 2025), 
we applied two complementary processing pipelines for creation-side and consumption-side analyses. To mitigate confounding~\cite{pearl2009causality}, we employed matching~\cite{matching} analysis. Methodological details and robustness checks for all matching analyses are provided in the Methods and Supplementary Information~\ref{SI:matching} (SI~\ref{SI:matching}), respectively.

\vspace{+5pt}
\noindent\textbf{AIGC creators produce higher volumes of content with overall engagement returns comparable to HGC creators.
}\\
\noindent Using 2,497 matched AIGC and HGC creators, we compared their content creation behaviour. We plotted the complementary cumulative distribution function (CCDF) of video creation volume for the matched creators. As shown in the left panel of Fig.~\ref{fig:results_1}(c), the CCDF curve for AIGC creators lies consistently above that of HGC creators (HL median difference $=4$, $p < 0.001$), indicating higher overall productivity. Further decomposition of AIGC creators’ output shows that this increase is driven primarily by AIGC video creation rather than HGC (SI \ref{SI:DecompositionCreator}).

We next assessed the engagement returns using total valid views, views exceeding a minimum watch-duration threshold, and full views, representing completed watches (Methods). The CCDFs of both metrics (Fig.~\ref{fig:results_1}(c) middle and right) are similar for AIGC and HGC creators, with Wilcoxon signed-rank tests indicating no meaningful differences (valid views: HL median difference$=0$, $p=0.003$; full views: HL median difference $\text{=}0$, $p=0.363$), 
indicating comparable engagement returns.
These results suggest that AIGC creators produce higher volumes of content with overall engagement returns comparable to HGC creators.

\vspace{+5pt}
\noindent\textbf{Consumers show weaker preferences for AIGC relative to HGC.}\\
Using 47,288 matched user-video interaction pairs, we compared consumer engagement preferences for AIGC versus HGC videos across three measures: valid-view rate (the fraction of interactions exceeding a minimum watch-duration threshold), full-view rate (the fraction of completed watches), and view duration (watch time per interaction). We reported average valid-view and full-view rates, and the distribution of view duration using box plots.
As shown in Fig.~\ref{fig:results_1}(d) (left panel), consumers display lower valid-view and full-view rates for AIGC videos than for HGC videos (valid-view rate: mean difference $=-0.076$, $p<0.001$; full-view rate: mean difference $=-0.027$, $p<0.001$). View duration follows the same pattern---consumers spend less time on AIGC videos (HL median difference $=-0.213s$, $p<0.001$; right panel of Fig.~\ref{fig:results_1}(d)). 
\sth{
}
These consistent underperformances suggest that consumers show weaker engagement preferences for AIGC videos.

\vspace{5pt}
\noindent\textbf{Scale-over-Preference dynamics of AIGC.}\\
The behavioral results reveal a clear SoP dynamic for AIGC: AIGC creators achieve aggregate engagement comparable to HGC creators through high-volume production, despite a marked consumer preference for HGC. This reflects an ecological tension in the content ecosystem, characterized by an asymmetry between AIGC supply scale and consumer preference. To quantify the tension, we defined a \textit{SoP index} as 
$ {SoPI} = \ln\!\left({S}/{P}\right),  S > 0,\; P > 0,$
where $S$ and $P$ denote AIGC’s relative supply scale and relative consumer preference at a given time, respectively, both measured relative to HGC. $\textit{SoPI}$ increases with larger $S$ and smaller $P$, corresponding to a larger tension. 
Fig.~\ref{fig:results_1}(e) visualizes the daily \text{SoP} index (\textit{i.e.}, \textit{SoPI} computed at each day) over the two-month observation period (April~1–May~31,~2025). 
All results cluster in a region of low relative consumer preference ($P \approx$ \text{0.30–0.40}) and higher relative supply scale ($S \approx \text{0.55–0.70}$), with most SoPI exceeding $\ln(1.5)$, suggesting a persistent tension between AIGC supply and consumer preference over time.

\subsection{Algorithmic Content Distribution Mechanism Moderates the Scale-over-Preference Dynamics of AIGC}

Algorithmic distribution mechanisms constitute the primary interface between creators and consumers, governing exposure allocation and shaping downstream outcomes. We next investigated how this mechanism responds to the rise of AIGC, focusing on exposure patterns and their subsequent effects on both stakeholders.

\begin{figure}[H]
    \centering
    \includegraphics[width=1\linewidth]{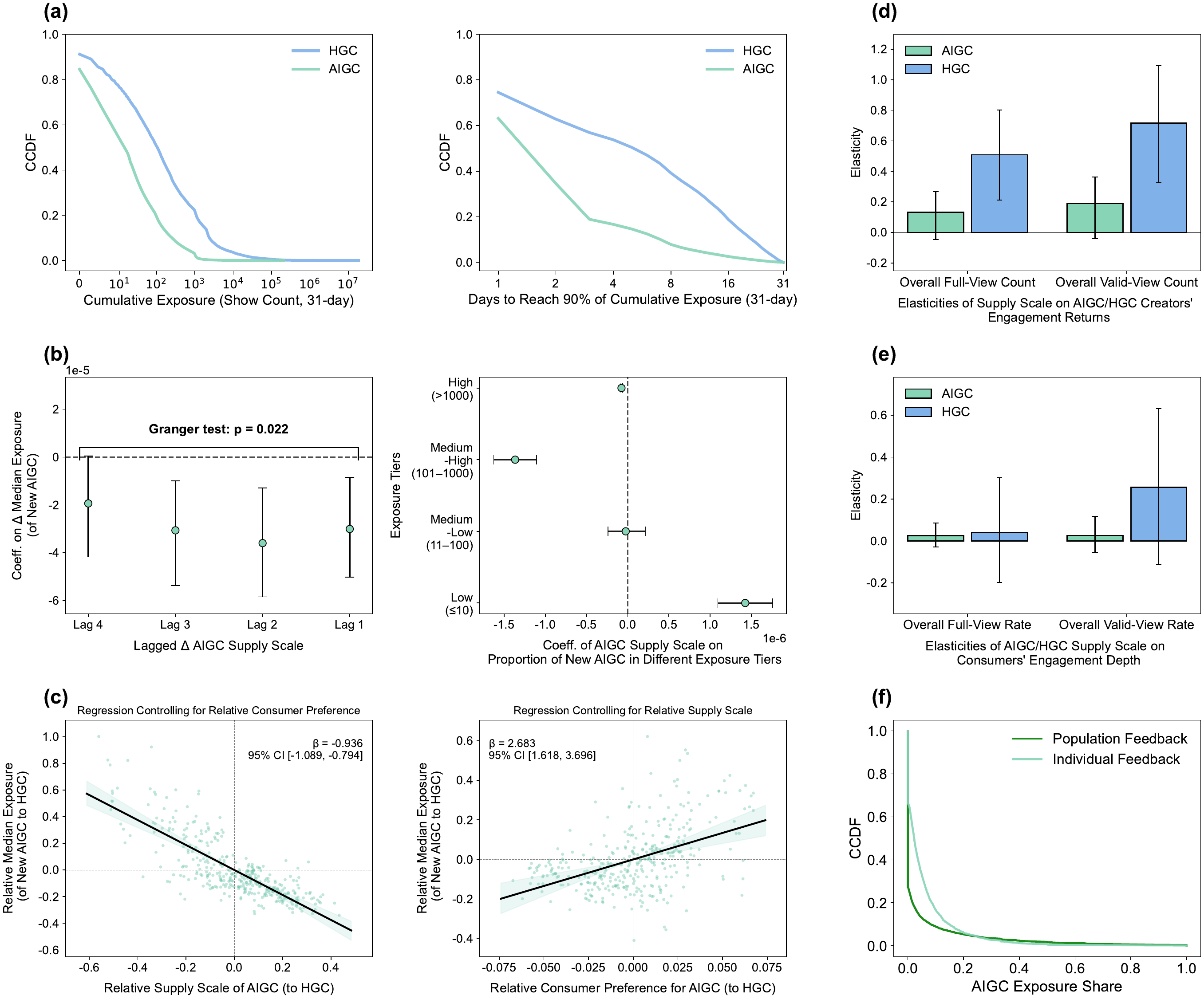}
   \caption{\textbf{
   Algorithmic content distribution mechanisms moderate the scale-over-preference dynamic.
   } 
\textbf{(a) Algorithmic exposure disadvantage and compressed lifecycle of AIGC.} CCDFs for matched video pairs ($n = 178,854$ pairs) show that AIGC receives lower cumulative exposure than HGC (left) and exhibits a more compressed exposure lifecycle (right; days to reach 90\% of 31-day exposure), consistent with weaker engagement preference for AIGC.
\textbf{(b) Algorithmic exposure response to AIGC supply expansion.} Granger analysis (left) with all negative lag coefficients indicates that increases in daily AIGC supply precede reductions in the median exposure of newly uploaded AIGC ($p = 0.022$; joint F-test). 
Regressions across exposure tiers (right) show that higher supply is associated with a downward shift in the exposure distribution: 
a positive association in the lowest tier ($\leq$10 views) and \sth{negative} or insignificant associations in higher tiers (11–100, 101–1000, $>$1000 views).
\textbf{(c) Algorithmic exposure response under varying SoP dynamics}. 
Dynamic regressions quantify exposure responses to rising tension between scale and preference. Conditional on relative preference for AIGC, AIGC exposure declines with increasing relative supply ($\beta$ = $-$0.936, 95\% CI = [$-$1.089, $-$0.794]; left).
Conditional on AIGC relative supply, AIGC exposure decreases with lower relative preference ($\beta$ = 2.683, 95\% CI = [1.618, 3.696]; right). Both
indicate the algorithm's moderation of AIGC visibility as SoP intensifies. 
\textbf{(d) Creator outcomes under algorithmic adjustment.} Log-log elasticities show that the \sth{overall}
creator engagement return elasticity with respect to AIGC supply is substantially smaller than for HGC, indicating \sth{smaller}
marginal returns to AIGC expansion (\sth{$SOPI$} $\uparrow$). Bars denote coefficients and error bars 95\%.
\textbf{(e) Consumer outcomes under algorithmic adjustment.} Log-log regression elasticities of \sth{overall} consumer engagement depth remain near zero as AIGC supply increases, indicating that overall consumer engagement does not decline under algorithmic moderation of AIGC expansion (\sth{$SOPI$} $\uparrow$).
Bars denote coefficients and error bars represent 95\% CIs,
and HGC results are reported for reference.
\textbf{(f) Heterogeneous responses across algorithms.} CCDFs of AIGC exposure ratios under two mechanisms show a leftward shift for the population feedback-driven algorithm relative to the individual feedback-driven algorithm, indicating that algorithm design differentially moderates the SoP dynamic.
}
    \label{fig:results_2}
\end{figure}

\vspace{5pt}
\noindent \textbf{Algorithmic content distribution mechanism assigns lower exposure to AIGC than to HGC.}\\
Using the dataset for consumption pattern analysis in Section~\ref{sec:sop}, we first examined algorithmic exposure over a 31-day post-upload window. To enable a fair comparison, we matched AIGC and HGC videos on factors like creator characteristics and content categories, yielding 178,854 matched pairs (robustness checks in SI~\ref{SI:matachingSensitivity}). We compared exposure along two dimensions: cumulative exposure (total show count) and exposure lifecycle (days to reach 90\% of cumulative exposure). As shown in Fig.~\ref{fig:results_2}(a), AIGC videos receive lower cumulative exposure (HL median difference = $-$59,  $p<0.001$ ) and reach 90\% exposure more quickly (HL median difference = $-$2 days, $p<0.001$) than matched HGC videos, indicating a more compressed exposure lifecycle. These results suggest that the algorithmic distribution mechanism allocates less and shorter exposure to AIGC, aligning with observed consumer preferences.

\vspace{5pt}
\noindent\textbf{Algorithmic content distribution mechanism exhibits a negative exposure response to AIGC supply expansion.} \\
We next examined how algorithmic distribution mechanisms respond to changes in AIGC supply using daily aggregated exposure data (June 2024–May 2025). Granger causality~\cite{Granger} analyses show that the lagged coefficients of AIGC supply on 31-day cumulative exposure are consistently negative (Fig.~\ref{fig:results_2}(b), left), indicating that increases in AIGC supply Granger-cause reductions in the cumulative exposure of newly uploaded AIGC videos (joint F-test: $F = 3.322, p = 0.022$). No significant Granger-causal relationship is observed for HGC videos, and the results are robust to alternative lag specifications and reverse-ordering tests (SI~\ref{SI:Granger}). 
\sth{Consistent with this pattern, higher AIGC supply has a positive association with the share of newly uploaded AIGC videos receiving very low exposure ($\le$10), but negative or insignificant associations with higher exposure tiers (Fig. 2(b), right).}
These findings suggest that algorithmic distribution mechanisms systematically downscale AIGC exposure as supply expands, helping to moderate scale-over-preference dynamics.

\vspace{5pt}
\noindent\textbf{Algorithmic content distribution mechanism moderates the scale-over-preference dynamics of AIGC.
}\\
We next examined how algorithmic distribution mechanisms respond to the varying SoP dynamics.
Note that the SoP index arises when relative AIGC supply ($S$) expands, or relative consumer preference ($P$) weakens.
We therefore regressed AIGC exposure on $S$ and $P$ to assess how exposure allocation changes as this tension intensifies. 
As shown in Fig.~\ref{fig:results_2}(c), conditional on consumer preference, AIGC exposure decreases with relative supply ($\beta=-0.936$, 95\% CI = [$-$1.089,$-$0.794]),
whereas conditional on supply, exposure increases with consumer preference ($\beta=2.683$, 95\% CI = [1.618, 3.696]), indicating that AIGC exposure was systematically reduced as the SoP index increased.

We also examined how this exposure adjustment translates into outcomes for creators and consumers as AIGC supply expands (\sth{$SOPI$} $\uparrow$). 
Creator outcomes are measured by aggregate engagement received, and consumer outcomes by engagement depth, defined as the average valid-view and full-view rates within 31 days after upload (see Method for definition details). 
\sth{
The log-log regression analyses in Fig.~\ref{fig:results_2}(d) show that the elasticity of aggregate engagement return with respect to AIGC supply is substantially smaller than that for HGC, indicating smaller marginal returns for AIGC creators with supply expansion.
In contrast, consumer engagement depth remains stable as AIGC supply increases (near-zero elasticities, Fig.~\ref{fig:results_2}(e)), suggesting that AIGC expansion has a negligible impact on overall consumer experience. 
The results show that the algorithmic distribution mechanism moderates the SoP dynamics via exposure adjustment, constraining the aggregate gains of AIGC creators while stabilizing consumer engagement depth.
}

Lastly, we examined how two distribution mechanisms (population feedback–driven and individual feedback–driven) moderate AIGC exposure by comparing their exposure for matched consumers ($n=4,735$ pairs\sth{; see Method for details}). The CCDF of the AIGC exposure ratio under the population feedback–driven algorithm is systematically shifted to the left relative to the individual feedback–driven algorithm (Fig.~\ref{fig:results_2}(f)), indicating stronger suppression of AIGC visibility when exposure is guided by population-level feedback. 
This contrast suggests that the SoP dynamic 
can be mediated by algorithmic design, rather than determined solely by supply and preference conditions.


\section{Discussion}

The rapid integration of AIGC is transforming digital content production into a hybrid human–AI ecosystem with sharply expanding AIGC supply. We identify a scale-over-preference dynamic: AIGC creators achieve aggregate engagement comparable to HGC mainly through high-volume production, despite weaker consumer preference for AIGC. This pattern reflects a supply-driven dissemination mode enabled by generative AI, in which visibility relies more on scale expansion than on preference-aligned creation, increasing the misalignment between supply and preference signals.

However, platform algorithms do not simply amplify this tendency. Distribution mechanisms partially attenuate the exposure of newly uploaded AIGC as SoP intensifies, yielding lower engagement–supply elasticity for AIGC than for HGC and 
\sth{less-than-proportional}
returns to scale-based creation. This moderation bounds the downstream impact of supply–preference misalignment while preserving overall user engagement. We further observe heterogeneity across algorithms, with population feedback–driven designs showing stronger mediating effects than individual feedback–driven ones.

These findings suggest that algorithmic distribution can mediate structural tensions introduced by large-scale AIGC. By documenting these dynamics and the differential responses of algorithmic designs, the study informs platform governance and highlights the need for AIGC-sensitive distribution strategies, potentially complemented by broader governance mechanisms, to constrain the accumulation of low-preference content as AIGC supply continues to grow.

\section{Method}

\subsection{Platform}
Our analysis is conducted on data from the Local Life Channel of a leading short-video platform in China (with over 400 million daily active users), where content exposure is primarily driven by algorithmic recommendations~\cite{kuairec,ncf,kruse2025design,wu2022survey} based on user–content interactions~\cite{kuairec}. 
On this platform, creators can upload content at high frequency and leverage built-in AIGC tools to assist video creation, enabling AIGC-based production to scale at substantially lower marginal effort than HGC and thereby amplifying variation in content supply across creators.
User engagement signals, such as valid views and full views, are recorded at scale and serve both as measures of content performance and as feedback signals for subsequent content distribution. These platform characteristics provide a suitable setting for studying AIGC's behavioral and distributional implications.

\subsection{Data}

\noindent\textbf{Identification of AIGC and HGC.}
AIGC and HGC are distinguished using platform-generated metadata labels embedded when videos are created with official AIGC tools. Videos carrying this label are treated as AIGC; all others as HGC. Potential misclassification from external AI tools is assessed using Sightengine~\cite{SightEngine}. Validation indicates that the unlabeled set is predominantly HGC, supporting the reliability of platform labels (SI-\ref{SI:aigc-label}).

\vspace{+5pt}
\noindent \textbf{Data  collection.} 
Data were collected from the platform's Local Life Channel spanning June 2024–May 2025.
To reduce confounding from heterogeneous browsing paths, analyses were restricted to impressions delivered through the main feed, where exposure is determined by algorithmic recommendation and user scrolling.

\vspace{+5pt}
\noindent \textbf{Data processing.}
Different experiments draw on different temporal subsets according to their objectives:

\vspace{+5pt}
\noindent  $\bullet$ \textit{For behavioral impact analyses.} 
Creator- and consumer-side behavior impact analyses use data from 1 April–31 May 2025. Creators are classified using a three-period design: pre-period (before 18 March), adoption period (18–30 March), and observation period (1 April–31 May). Creators with no prior AIGC usage who first adopt AIGC during the adoption period and upload $\geq$1 AIGC video in the observation period are defined as AIGC creators; active creators who do not use AIGC during all periods are defined as HGC creators, yielding 10,763 AIGC creators and 376,713 HGC creators.
All videos uploaded in the observation period are tracked for 31 days post-upload.
For consumption-side analyses, we construct a balanced sample of 0.8M AIGC and 0.8M HGC videos from the same window and randomly sample 24M interactions (12M for each type).



\vspace{+5pt}
\noindent  $\bullet$ \textit{For algorithmic response analyses.} Analyses of algorithmic responses are conducted under two settings. (1) In the non-dynamic setting, 
we use the same consumption-side sample with all exposure records for distributional difference examination.
(2) In the dynamic setting, we use the full one-year dataset (June 2024–May 2025) to capture how the platform's algorithmic distribution mechanism reacts 
under varying scale-over-preference conditions, covering all exposure records for approximately 118 million videos.

\subsection{Evaluation Metrics}
We use a multi-dimensional set of metrics covering creator-side, consumer-side, and distributional outcomes, derived from the collected interaction/exposure logs.


\vspace{+3pt}
\noindent\textbf{Creator-side metrics.}
\textit{Creation volume} is the number of videos uploaded by a creator during the observation period. Creator \textit{engagement return} is measured by aggregating \textit{valid view} and \textit{full view} counts received by all videos uploaded by the creator. Following platform definitions, 
a valid view is recorded when a video $<$7s is watched in full, or a video $\geq$7s accumulates $\geq$7s playback. A full view denotes completion of the video.


\vspace{+3pt}
\noindent\textbf{Consumer-side metrics.}
Consumer \textit{engagement depth} is measured by average \textit{view duration}, \textit{valid-view rate} (valid views divided by total views), and \textit{full-view rate} (full views divided by total views), capturing attention from initial to sustained viewing.

\vspace{+3pt}
\noindent\textbf{Distributional metrics.}
Algorithmic exposure is measured by \textit{show count} (impressions delivered by the recommender). We additionally compute \textit{time-to-90\% exposure}—days required to reach 90\% of a video’s 31-day cumulative shows; shorter values indicate a more front-loaded exposure lifecycle.


\vspace{+3pt}
\noindent\textit{Evaluation horizon and data consistency.}
All metrics are calculated on a 31-day horizon after upload, covering $>$80\% of total activity. Records are deduplicated and filtered for abnormal or non-human traffic prior to computation.


\subsection{Analytical Framework}

The study employs two analytical streams—behavioral analyses and algorithmic response analyses.

\subsubsection{Behavioral Analyses}  To examine the behavioral effects of AIGC on creation and consumption, we employ matching-based analyses to mitigate potential confounding. Because the creation-side and consumption-side analyses focus on different units, we apply distinct matching strategies as described below.

\vspace{+5pt}
\noindent\textbf{Matching for creation-side analysis.}
To estimate effects on creator behavior, we match AIGC creators with comparable HGC creators based on pre-treatment behavioral profiles and metadata characteristics. Pre-treatment covariates are derived from the period before adoption and include creators’ 14-/60-day upload volumes, received play counts, and total watch time. Metadata controls comprise follower count, content category, city, and brand tier. Exact matching is applied to metadata (except follower count), and nearest neighbor matching to behavioral features and follower count using Euclidean distance on z-score standardized data (z-score: $z_i=\frac{x_i - mean}{std}$). Creators are considered matched only when all criteria are satisfied.

\vspace{+5pt}
\noindent\textbf{Matching for consumption-side analysis.}
Consumption-side analyses are conducted at the interaction level, with matching applied to users, videos, and temporal context.
On the user side, we control recent behavior signals measured prior to each interaction, including 14-/60-day AIGC/HGC play counts and watch time, as well as gender, age, and city.
On the video side, we control creator-related characteristics measured prior to video upload—14-/60-day upload volumes, received play counts and watch time, creator category, city, and follower count. We additionally control video upload date and interaction time of day (morning, noon, afternoon, night, late night) to account for temporal heterogeneity.
Consistent with the creation-side procedure, continuous features are matched using the same nearest neighbor matching, and discrete features are matched using exact matching.

\vspace{+5pt}
\subsubsection{ \textit{Algorithmic Response Analyses}} 
Algorithmic response analyses consist of 
(i) matching-based analyses to compare AIGC and HGC exposure within fixed periods \sth{and compare different algorithms' effects},
and (ii) regression-based time-series analyses examining how system exposure responds to changes in the scale-over-preference dynamic.  

\vspace{+5pt}
\noindent\textbf{Matching-based comparisons.} 
The comparisons first contrast the exposure received by AIGC and HGC within the same time window, aiming to estimate algorithmic distributional differences under comparable conditions. We follow the matching procedures described for the consumption-side behavioral analysis, focusing on video-level matching because the unit of analysis is the video without conditioning on specific users. Specifically, we match AIGC and HGC videos based on creator-related characteristics prior to the video upload—14-day and 60-day upload volumes, play counts received, watch time received, creator category, city, and follower count—as well as video upload date.

Matching analysis is also used to evaluate the moderating effects of different algorithmic designs (individual feedback–driven vs. population feedback–driven) on AIGC exposure. To isolate these effects, we match consumers across the two algorithm groups based on their profiles, including recent consumption preferences (14- and 60-day AIGC/HGC play counts and watch time) and demographic characteristics (city, gender, and age). The item pool is identical for both algorithms and is therefore not included as a control.


\vspace{+10pt}
\noindent\textbf{Regression-based time-series analyses.}
To investigate how the algorithmic content distribution mechanism responds to the SoP dynamic, we conduct two complementary analyses: Granger causality tests~\cite{Granger_orginal,Granger} for directional assessment and Dynamic OLS (DOLS)~\cite{stock1993simple} for estimating long-run associations.

\vspace{+5pt}
\noindent (1) Granger causality analysis. We aggregate the AIGC supply scale and the median exposure of newly uploaded AIGC at day $t$. Let $E_{t}$ denote the median exposure of newly uploaded AIGC videos. Directional dependence is examined using a regression F-test that compares an unrestricted full model against a restricted model. We specify the unrestricted full model as:


\begin{equation}
\Delta E_{t} = \alpha + \sum_{i=1}^{p}\beta_{i}\Delta E_{t-i} + \sum_{i=1}^{p}\gamma_{i}\Delta S_{t-i} + \sum_{i=1}^{p}\delta_{i}\Delta S_{t-i}^{\prime} + \lambda D_{t} + \epsilon_{t}
\end{equation}

where $\Delta$ denotes the first-difference operator (i.e., $\Delta X_{t} = X_{t} - X_{t-1}$, representing the difference between two consecutive days). 
Here, $S_{t}$ is the absolute AIGC supply scale at day $t$, serving as the focal explanatory variable. $S_{t}^{\prime}$ represents the HGC supply scale; its lagged terms ($S_{t-i}^{\prime}$) are included as controls to account for volume correlations and isolate AIGC-specific dynamics.
Additionally, $D_{t}$ denotes a binary dummy variable for weekdays and holidays to account for temporal heterogeneity, and $\epsilon_{t}$ is the error term. 

To test for Granger causality, we perform a joint F-test comparing this full model to a restricted model where the coefficients of the lagged focal supply variables are constrained to zero ($H_{0}: \gamma_{1} = \gamma_{2} = \dots = \gamma_{p} = 0$). 
Rejection of $H_{0}$ (p=4 in our experiments) indicates that incorporating the changes in the historical supply scale of AIGC significantly improves the prediction of its future exposure changes.


\vspace{+5pt}
\noindent(2) Dynamic OLS (DOLS) analysis.
To estimate long-run associations between SoP-related factors and algorithmic exposure, we used  DOLS models, which incorporate leads and lags of differenced regressors to address potential endogeneity and serial correlation. For each analysis target variable $Y_t$, we estimate:
\begin{equation}
    Y_t = \alpha + \beta X_t + \theta C_{t} + \sum_{j=-p}^{p} \gamma_{j} \Delta X_{t-j} + \epsilon_t + \delta D_t,
\end{equation}
where $X_t$ is the focal explanatory variable, $C_t$ denotes the controlled variables, $\Delta X_{t-j}$ are lead–lag adjustments, $\epsilon_t$ denotes the error term, $D_t$ denotes the weekday and statutory-holiday indicator (binary) to control for temporal heterogeneity. The coefficient $\beta$ captures the association between SoP-related factors and algorithmic exposure.

\textit{Operationalizations for analyses.} We conduct five analyses to operationalize algorithmic response at different levels. (a) links AIGC supply scale to the entry of new AIGC items into different exposure tiers, representing the initial visibility allocation. (b–c) relate AIGC exposure to two dimensions of the scale-over-preference condition: (b) relative supply scale and (c) relative preference. (d–e) relate the AIGC supply scale to downstream outcomes, including (d) creators’ overall engagement returns and (e) consumers’ overall engagement depth. The corresponding specifications are:

\noindent(a) New AIGC entry into exposure tiers.
\begin{equation}
Y_t = \pi^k_t, \quad X_t = S_t^{AIGC},
\end{equation}
where $\pi^k_t$ is the proportion of new AIGC entering exposure tier $k$, and $S_t^{AIGC}$ denotes the absolute supply scale of AIGC. Besides, the absolute supply scale of HGC, denoted by $S_t^{HGC}$, is considered as the controlled variable, i.e., $C_t = S_t^{HGC}$.

\vspace{0.5em}
\noindent(b)\&(c) AIGC relative supply / consumer's relative preference on AIGC $\rightarrow$ AIGC relative exposure.
\begin{equation}
Y_t = E_t, \quad X_t = \{S_t, P_t\},
\end{equation}
where $E_t$ denotes the relative median exposure of AIGC to HGC on day $t$, and $P_t$ represents the relative consumer preference for AIGC on day $t$, measured by valid-view rates. Here, as $S_t$ (the relative supply scale of AIGC to HGC) and $P_t$ jointly determine the SoP, we analyzed them together.

\vspace{0.5em}
\noindent(d) AIGC absolute supply scale $\rightarrow$ creator's overall engagement return:
\begin{equation}
Y_t = \ln(ER_t^{AIGC}), \quad X_t = \ln(S_t^{AIGC}),
\end{equation}
where $ER_t^{AIGC}$ represents the overall engagement return for AIGC creators, calculated on day $t$. $S_t^{HGC}$ is treated as the controlled variable. For reference, we also performed the ``HGC absolute supply scale $\rightarrow$ creator's overall engagement return'' analysis, setting $X_t = \ln(S_t^{HGC})$ and $Y_t=\ln(ER_t^{HGC})$ as HGC creators' overall engagement returns, while controlling $S_t^{AIGC}$.

\vspace{0.5em}
\noindent(e) Absolute supply scale $\rightarrow$ consumer's overall engagement depth:
\begin{equation}
Y_t = \ln(ED_t), \quad X_t = \{\ln(S_t^{AIGC}), \ln(S_t^{HGC})\},
\end{equation}
where $ED_t$ represents the overall consumer engagement depth at day $t$. Notably, since consumer depth is jointly determined by AIGC and HGC, we involved both $S_t^{AIGC}$ and $S_t^{HGC}$ for the regression.

\bibliography{sn-bibliography}


\clearpage
\section*{Supplementary Information}
\setcounter{NAT@ctr}{0}
\setcounter{section}{0}
\setcounter{page}{1}
\setcounter{table}{0}
\setcounter{figure}{0}
\renewcommand{\tablename}{Supplementary Table}
\renewcommand{\figurename}{Supplementary Fig.}

\renewcommand{\thesection}{\arabic{section}}




\section{Supplementary Information on Data}

\subsection{Validation of Platform AIGC Labels}
\label{SI:aigc-label}

In our main analysis, AIGC and HGC are distinguished using platform-generated metadata labels.
A potential threat to validity is that creators might use external AI tools without triggering the platform's built-in AIGC label, leading to the misclassification of some AIGC as HGC. 
To assess this potential misclassification and evaluate the reliability of the unlabeled set as a proxy for HGC, we conducted an external validation using the Sightengine AI's~\cite{SightEngine} AI-generated video detection API.

We randomly sampled 114 platform-labeled AIGC videos and 339 unlabeled videos (treated as HGC). The API identified 69.3\% (79/114) of the platform-labeled AIGC videos as AI-generated, verifying the effectiveness of the detector. In contrast, only 24.2\% (82/339) of the unlabeled videos were flagged. This indicates that the vast majority of the unlabeled videos are indeed non-AIGC, thereby supporting the reliability of using platform labels for our large-scale classification.

\subsection{Duration of
 Videos}
 \label{SI:video-length}
 Supplementary Table~\ref{SI.Table:video.duration} shows the statistics of Video Duration (Seconds), presenting the physical duration of the videos in our sample after deduplicating the interaction logs. The duration is measured in seconds. The results show that HGC is generally longer on average.

 \begin{table}[h]
\centering
\caption{\textbf{Statistics of video duration (seconds).} 
This table presents the physical duration of the videos in our sample after deduplicating the interaction logs. The duration is measured in seconds.}
\begin{tabular}{l|cccccc}
\hline
\textbf{Content Type}  & \textbf{Mean} & \textbf{Std. Dev.} & \textbf{P25} & \textbf{Median} & \textbf{P75} \\ \hline
\textbf{AIGC Video}          & 20.69             & 7.05               & 17.80            & 20.11               & 22.77            \\
\textbf{HGC Video}           & 40.75             & 53.42              & 15.97            & 25.80               & 49.53            \\ \hline
\end{tabular}
\label{SI.Table:video.duration}
\end{table}




\section{Matching}
\label{SI:matching}





To mitigate confounding issues, we perform matching in our analyses, including creation-side comparison, consumption-side comparison, algorithmic distribution comparison, and algorithmic comparison. We further provide detailed descriptions of the matching method and conduct balance and sensitivity analyses to assess the robustness and reliability of the results.

\subsection{Matching for algorithmic mechanism comparison}
We aim to evaluate how algorithmic designs moderate AIGC exposure by comparing outcomes under population feedback-driven versus individual feedback-driven mechanisms. To isolate the pure algorithmic effect from baseline consumer habits, we matched consumers across the two algorithm groups based strictly on their historical profiles. We controlled for recent consumption preferences (14- and 60-day AIGC/HGC play counts and watch time) alongside demographic characteristics (city, gender, and age).
From an initial pre-match pool of 10,000 consumers under the population feedback-driven algorithm and 1,000,000 consumers under the individual feedback-driven algorithm, our procedure yielded 4,735 matched consumer pairs.



\subsection{Matching Conditions and Distance Metrics}
To operationalize the matching, we established a rigorous set of rules combining exact matching for discrete attributes with nearest-neighbor distance minimization for continuous variables. 

\noindent \textbf{Exact matching for categorical features:} 
For categorical and discrete metadata, we enforced a strict exact matching condition. This applies to features such as city, gender, predefined age segments (utilizing the platform's native demographic brackets), creator's content category, brand tier, video upload dates, and interaction time of day. In our algorithm, these discrete features act as absolute hard filters: if a potential control unit does not perfectly align with the treatment unit across all relevant categorical dimensions, the pair is immediately disqualified.

\noindent \textbf{Distance calculation for numerical features:}
For continuous metrics (e.g., 14-/60-day upload volumes, play counts, total watch time, and follower counts), the natural scales and variances differ drastically. To prevent variables with larger magnitudes from dominating the distance calculation, we first standardized all continuous features using $z$-score normalization:
\begin{equation*}
    z_{i,k} = \frac{x_{i,k} - \mu_{k}}{\sigma_{k}}
\end{equation*}
where $x_{i,k}$ is the raw value of feature $k$ for unit $i$, and $\mu_{k}$ and $\sigma_{k}$ are the mean and standard deviation of feature $k$ across the respective candidate pool. Following normalization, the dissimilarity between a treatment unit $t$ and a candidate control unit $c$ was quantified using the Euclidean distance across all $K$ standardized continuous features:
\begin{equation*}
  d(t, c) = \sqrt{\sum_{k=1}^{K} (z_{t,k} - z_{c,k})^2}.  
\end{equation*}

\noindent \textbf{Matching algorithm:} 
Subject to the exact matching constraints on discrete features, we implemented a 1:1 Nearest Neighbor (NN) matching algorithm with replacement~\cite{abadie2006large}. For each unit in the treatment group, the algorithm searches the eligible control pool and selects the single unit that minimizes the Euclidean distance $d(t, c)$ (with the distance defined as infinite for units that fail to meet categorical matching constraints).
Allowing matching with replacement significantly reduces bias and improves the overall quality of the matches, as it permits a highly representative control unit to serve as the optimal counterfactual for multiple treatment units.

\begin{table}[ht]
\label{tab:video_duration}
\caption{\textbf{Covariate balance test for creation-side matching (numerical features)}. 
To accurately assess changes in creator behavior, this matching ensures that AIGC creators and HGC creators had comparable baseline productivity (e.g., historical upload volumes) and influence (e.g., follower count, historical play counts) prior to AIGC adoption.
The "AIGC Creator" and "HGC Creator" columns report the mean values for the respective groups. SMD stands for Standardized Mean Difference. VR stands for Variance Ratio.}
\begin{tabular}{c|c|cccc}
\hline
Group                                                                      & \textbf{Feature}                      & 
\begin{tabular}[c]{@{}c@{}}\textbf{AIGC}\\ \textbf{Creator}\end{tabular}

& \begin{tabular}[c]{@{}c@{}}\textbf{HGC}\\ \textbf{Creator}\end{tabular} & \textbf{SMD} & \textbf{VR} \\ \hline
\multirow{7}{*}{\begin{tabular}[c]{@{}c@{}}Before\\ Matching\end{tabular}} & Creator Follower Count & 31.33              & 32.58            & -0.0026      & 0.57        \\
                                                                           & Creator Video Uploads (14 days)       & 19.96              & 4.98             & 0.1401       & 19.31       \\
                                                                           & Creator Video Uploads (60 days)       & 24.64              & 10.57            & 0.0624       & 17.61       \\
                                                                           & Creator Total Play Hours (14 days)    & 21.97              & 6.92             & 0.0801       & 1.31        \\
                                                                           & Creator Total Play Hours (60 days)          & 35.30               & 19.71            & 0.0412       & 1.36        \\
                                                                           & Creator Total Play Count (14 days)          & 11044.63           & 1670.50           & 0.1927       & 5.72        \\
                                                                           & Creator Total Play Count (60 days)          & 15862.46           & 5104.32          & 0.1064       & 4.41        \\ \hline
\multirow{7}{*}{\begin{tabular}[c]{@{}c@{}}After\\ Matching\end{tabular}}  & Creator Follower Count & 14.94              & 14.61            & 0.0030         & 0.84        \\
                                                                           & Creator Video Uploads (14 days)       & 7.21               & 5.99             & 0.0608       & 0.97        \\
                                                                           & Creator Video Uploads (60 days)       & 8.57               & 8.43             & 0.0063       & 1.22        \\
                                                                           & Creator Total Play Hours (14 days)          & 5.48               & 5.57             & -0.0028      & 1.00           \\
                                                                           & Creator Total Play Hours (60 days)          & 8.53               & 8.67             & -0.0033      & 1.01        \\
                                                                           & Creator Total Play Count (14 days)          & 2471.61            & 2146.41          & 0.0294       & 1.52        \\
                                                                           & Creator Total Play Count (60 days)          & 3517.02            & 3297.52          & 0.0161       & 1.33        \\ \hline
\end{tabular}
\label{tab:balance_creator}
\end{table}

\subsection{Matching Balance Test}
To evaluate matching quality, we assessed covariate balance using the Standardized Mean Difference (SMD) and Variance Ratio (VR). Following Rubin's recommendations~\cite{rosenbaum1985constructing,rubin2001using}, a covariate is considered adequately balanced if $|SMD| < 0.1$ and $VR \in [0.5, 2.0]$. We report these metrics exclusively for numerical features, as our exact matching protocol inherently guarantees perfect balance for all categorical variables.

Supplementary Tables~\ref{tab:balance_creator}, \ref{tab:balance_interaction}, \ref{tab:balance_video}, and \ref{tab:balance_algorithm} present the detailed diagnostics for our four matching tasks. While substantial baseline disparities existed prior to matching, our procedures effectively eliminated these selection biases. Post-matching, the SMDs for all continuous covariates converged strictly below the 0.1 threshold, and all VRs stabilized within the $[0.5, 2.0]$ bounds (with the vast majority approaching 1.0). 
These results suggest that the treatment and control groups are highly comparable after matching, which helps support more reliable downstream estimations.

\begin{table}[ht]
\caption{\textbf{Covariate balance test for consumption-side matching (numerical features).} 
To accurately measure consumer engagement, this interaction-level matching controls for both the historical viewing preferences of the consumer and the historical performance of the viewed video's creator. 
The "AIGC" and "HGC" columns report the mean values for the respective interaction groups. SMD stands for Standardized Mean Difference. VR stands for Variance Ratio.}
\begin{tabular}{c|c|cccc}
\hline
\textbf{Group}                                                                        & \textbf{Feature}                   & \textbf{AIGC} & \textbf{HGC} & \textbf{SMD} & \textbf{VR} \\ \hline
\multirow{15}{*}{\begin{tabular}[c]{@{}c@{}}Before \\ Matching\end{tabular}} 
                                                                             & Creator Follower Count        & 935.20              & 1490.10           & -0.0731      & 0.39        \\
                                                                             & Creator Video Uploads (14 days)    & 196.70              & 318.51           & -0.133       & 0.33        \\
                                                                             & Creator Video Uploads (60 days)    & 587.01             & 1141.83          & -0.1629      & 0.25        \\
                                                                             & Creator Total Play Hours (14 days) & 242.32             & 2177.65          & -0.4238      & 0.04        \\
                                                                             & Creator Total Play Hours (60 days) & 686.40              & 7295.51          & -0.4088      & 0.03        \\
                                                                             & Creator Total Play Count (14 days) & 150476.74          & 1374554.76       & -0.3991      & 0.04        \\
                                                                             & Creator Total Play Count (60 days) & 393599.19          & 4383287.51       & -0.3614      & 0.02        \\
                                                                             & Consumer AIGC Play Hours (14 days) & 3.05               & 0.86             & 0.1431       & 4.90         \\
                                                                             & Consumer HGC Play Hours (14 days)  & 8.31               & 5.78             & 0.0427       & 1.29        \\
                                                                             & Consumer AIGC Play Hours (60 days) & 11.83              & 6.05             & 0.0801       & 1.90         \\
                                                                             & Consumer HGC Play Hours (60 days)  & 35.88              & 22.74            & 0.0550        & 1.51        \\
                                                                             & Consumer AIGC Play Count (14 days) & 8599.95            & 3347.05          & 0.1177       & 2.12        \\
                                                                             & Consumer HGC Play Count (14 days)  & 27841.96           & 21671.71         & 0.0282       & 0.91        \\
                                                                             & Consumer AIGC Play Count (60 days) & 37366.81           & 20615.32         & 0.0702       & 1.71        \\
                                                                             & Consumer HGC Play Count (60 days)  & 114411.42          & 75261.79         & 0.0490        & 1.34        \\
 \hline
\multirow{15}{*}{\begin{tabular}[c]{@{}c@{}}After \\ Matching\end{tabular}}  & Creator Follower Count        & 3097.61            & 3210.83          & -0.0119      & 0.93        \\
                                                                             & Creator Video Uploads (14 days)    & 486.49             & 481.69           & 0.0043       & 0.94        \\
                                                                             & Creator Video Uploads (60 days)    & 1332.83            & 1391.94          & -0.0162      & 0.91        \\
                                                                             & Creator Total Play Hours (14 days) & 722.74             & 744.29           & -0.0098      & 0.97        \\
                                                                             & Creator Total Play Hours (60 days) & 1678.69            & 1844.06          & -0.0323      & 0.92        \\
                                                                             & Creator Total Play Count (14 days) & 450641.30           & 463073.87        & -0.0081      & 0.93        \\
                                                                             & Creator Total Play Count (60 days) & 964622.33          & 1048336.13       & -0.0265      & 0.87        \\
                                                                             & Consumer AIGC Play Hours (14 days) & 0.02               & 0.02             & 0.0465       & 1.04        \\
                                                                             & Consumer HGC Play Hours (14 days)  & 0.23               & 0.24             & -0.0396      & 1.02        \\
                                                                             & Consumer AIGC Play Hours (60 days) & 0.05               & 0.05             & 0.0234       & 0.93        \\
                                                                             & Consumer HGC Play Hours (60 days)  & 0.84               & 0.87             & -0.0517      & 0.96        \\
                                                                             & Consumer AIGC Play Count (14 days) & 27.14              & 25.18            & 0.0652       & 1.18        \\
                                                                             & Consumer HGC Play Count (14 days)  & 186.32             & 202.75           & -0.0981      & 0.89        \\
                                                                             & Consumer AIGC Play Count (60 days) & 81.00                 & 78.18            & 0.0308       & 1.12        \\
                                                                             & Consumer HGC Play Count (60 days)  & 663.60              & 720.23           & -0.0955      & 0.89        \\
 \hline
\end{tabular}
\label{tab:balance_interaction}
\end{table}

\begin{table}[]
\caption{
\textbf{Covariate balance test for algorithmic distribution comparison (numerical features)}. 
To fairly compare the algorithmic exposure allocated to different content, this video-level matching ensures that AIGC videos and HGC videos shared comparable pre-upload creator characteristics (e.g., historical upload volumes and play counts). 
The "AIGC Video" and "HGC Video" columns report the mean values for the respective groups. SMD stands for Standardized Mean Difference. VR stands for Variance Ratio.
}
\begin{tabular}{c|c|cccc}
\hline
Group                                                                      & \textbf{Feature}                   & \textbf{AIGC Videos} & \textbf{HGC Videos} & \textbf{SMD} & \textbf{VR} \\ \hline
\multirow{7}{*}{\begin{tabular}[c]{@{}c@{}}Before\\ Matching\end{tabular}} & Creator Follower Count        & 594.53             & 2314.72          & -0.2587      & 0.14        \\
                                                                           & Creator Video Uploads (14 days)    & 197.94             & 709.72           & -0.3679      & 0.08        \\
                                                                           & Creator Video Uploads (60 days)    & 591.63             & 2642.00             & -0.3754      & 0.05        \\
                                                                           & Creator Total Play Hours (14 days) & 178.12             & 982.94           & -0.3098      & 0.09        \\
                                                                           & Creator Total Play Hours (60 days) & 539.79             & 3887.15          & -0.3393      & 0.07        \\
                                                                           & Creator Total Play Count (14 days) & 104990.54          & 346783.10         & -0.2180       & 0.22        \\
                                                                           & Creator Total Play Count (60 days) & 309371.15          & 1317096.59       & -0.2708      & 0.13        \\
 \hline
\multirow{7}{*}{\begin{tabular}[c]{@{}c@{}}After\\ Matching\end{tabular}}  & Creator Follower Count        & 1802.20             & 1799.99          & 0.0003       & 1.00        \\
                                                                           & Creator Video Uploads (14 days)    & 413.53             & 408.68           & 0.0058       & 0.99        \\
                                                                           & Creator Video Uploads (60 days)    & 1245.69            & 1225.70           & 0.0072       & 1.01        \\
                                                                           & Creator Total Play Hours (14 days) & 497.65             & 501.19           & -0.0018      & 1.00        \\
                                                                           & Creator Total Play Hours (60 days) & 1488.18            & 1505.31          & -0.0025      & 1.00        \\
                                                                           & Creator Total Play Count (14 days) & 276955.85          & 275162.12        & 0.0015       & 1.01        \\
                                                                           & Creator Total Play Count (60 days) & 788405.14          & 783340.1         & 0.0015       & 1.01        \\
 \hline
\end{tabular}
\label{tab:balance_video}
\end{table}

\begin{table}[]
\centering
\caption{
\textbf{Covariate balance test for algorithmic mechanism comparison (numerical features)}. 
To robustly compare the two distribution mechanisms, this consumer-level matching ensures that consumers in both algorithm groups had comparable historical consumption behaviors (\textit{e.g.}, historical play hours and play counts). The "Indiv. Feed." (short for ``individual feedback") and "Pop. Feed." (short for ``population feedback") columns report the mean values for the respective consumer groups receiving the algorithm based on individual/population-level feedback. SMD stands for Standardized Mean Difference. VR stands for Variance Ratio.
}
\begin{tabular}{c|c|cccc}
\hline
\textbf{Group}                                                             & \textbf{Feature}                   & \textbf{Indiv. Feed.} & \textbf{Pop. Feed.} & \textbf{SMD} & \textbf{VR} \\ \hline
\multirow{8}{*}{\begin{tabular}[c]{@{}c@{}}Before\\ Matching\end{tabular}} & Consumer AIGC Play Hours (14 days) & 5.38               & 4.11             & 0.0358       & 0.40         \\
                                                                           & Consumer HGC Play Hours (14 days)  & 77.08              & 73.73            & 0.0153       & 1.09        \\
                                                                           & Consumer AIGC Play Count (14 days) & 4.84               & 4.54             & 0.0060        & 0.06        \\
                                                                           & Consumer HGC Play Count (14 days)  & 26.01              & 45.66            & -0.2454      & 0.29        \\
                                                                           & Consumer AIGC Play Hours (60 days) & 13.57              & 12.75            & 0.0061       & 0.20         \\
                                                                           & Consumer HGC Play Hours (60 days)  & 276.38             & 310.71           & -0.0552      & 0.53        \\
                                                                           & Consumer AIGC Play Count (60 days) & 11.7               & 14.49            & -0.0117      & 0.01        \\
                                                                           & Consumer HGC Play Count (60 days)  & 95.88              & 197.43           & -0.2869      & 0.13        \\ \hline
\multirow{8}{*}{\begin{tabular}[c]{@{}c@{}}After\\ Matching\end{tabular}}  & Consumer AIGC Play Hours (14 days) & 2.38               & 2.23             & 0.0317       & 1.49        \\
                                                                           & Consumer HGC Play Hours (14 days)  & 75.63              & 71.99            & 0.0351       & 1.03        \\
                                                                           & Consumer AIGC Play Count (14 days) & 2.83               & 2.62             & 0.0452       & 1.32        \\
                                                                           & Consumer HGC Play Count (14 days)  & 28.39              & 28.35            & 0.0012       & 1.05        \\
                                                                           & Consumer AIGC Play Hours (60 days) & 4.68               & 4.55             & 0.0117       & 1.41        \\
                                                                           & Consumer HGC Play Hours (60 days)  & 272.18             & 253.38           & 0.0609       & 1.26        \\
                                                                           & Consumer AIGC Play Count (60 days) & 5.80                & 5.56             & 0.0208       & 1.19        \\
                                                                           & Consumer HGC Play Count (60 days)  & 102.07             & 102.03           & 0.0003       & 1.15        \\ \hline
\end{tabular}
\label{tab:balance_algorithm}
\end{table}

\subsection{Sensitivity Analysis}
\label{SI:matachingSensitivity}


\subsubsection{Methodological Framework}

While our matching procedure ensures strict comparability across observed characteristics, observational estimates remain inherently susceptible to unobserved confounding. 
To systematically assess the robustness of our estimates against potential omitted variable bias, we implemented the sensitivity analysis framework introduced by Cinelli and Hazlett~\cite{cinelli2020making}. 
The core intuition of this framework is: rather than relying on the untestable assumption that no unobserved confounders exist, it quantifies exactly how influential such an unobserved confounder must be to overturn the results.

This quantification is achieved by reparameterizing the required strength of unobserved confounding in terms of partial variance explained ($R^2$).
Specifically, the baseline strength of the estimated impact is first captured by $R^2_{Y \sim D|\textbf{X}}$, representing the proportion of residual variance in the outcome uniquely explained by the treatment assignment ($D$) after controlling for all observed covariates ($X$). Building upon this baseline strength, the framework calculates the Robustness Value ($RV$). 
$RV_{q=1}$ defines the critical threshold: the minimum explanatory power a hypothetical unobserved confounder ($Z$) must share with both the treatment and the outcome to completely eliminate the estimated impact. 
Correspondingly, $RV_{q=1,\alpha=0.05}$ defines the threshold required to render the point estimate statistically insignificant at the 5\% level, which serves as a more conservative robustness bound than $RV_{q=1}$.

To assess whether an unobserved confounder could realistically reach the $RV$ threshold, this framework benchmarks it against the empirical strength of the observed covariates. 
Specifically, it calculates how strongly each known variable ($Z_{k}$) influences the treatment assignment ($R^2_{D \sim Z_{k}|\textbf{X}}$) and the outcome ($R^2_{Y \sim Z_{k}|\textbf{X},D}$). 
Because our models already incorporate highly predictive factors as confounders---such as historical engagement and creator influence---the maximum explanatory power among these known variables serves as a conservative empirical bound. If the calculated $RV$ substantially exceeds this bound, the results are deemed highly robust.
Consequently, if the calculated $RV$ is multiple times larger than these bounds ($R^2_{D \sim Z_{k}|\textbf{X}}$ and $R^2_{Y \sim Z_{k}|\textbf{X},D}$), overturning the results would require an unobserved confounder vastly more powerful than the observed primary drivers---a highly improbable scenario that could support the robustness of our estimates.

\begin{table}[t]
\centering
\caption{\textbf{Sensitivity analysis for creation-side behavioral outcomes.} \textit{Coef.} reports the regression coefficient of the treatment (creator's adoption of AIGC tools) for each respective outcome (creator behavioral metrics). $R^2_{Y \sim D | \mathbf{X}}$ denotes the proportion of residual outcome variance explained by the treatment. $RV_{q=1}$ and $RV_{q=1, \alpha=0.05}$ represent the robustness values required for an unobserved confounder to reduce the estimate to zero or render it statistically insignificant at the 5\% level, respectively. The reference columns report the maximum explanatory power of the observed covariates ($Z_k$) on the treatment ($R^2_{D \sim Z_k | \mathbf{X}}$) and the outcome ($R^2_{Y \sim Z_k | \mathbf{X}, D}$), identified by iterating through all candidates to establish an empirical benchmark.
*** $p < 0.001$, ** $p < 0.01$, * $p < 0.05$.}
\begin{tabular}{l|cccccc}
\toprule
\begin{tabular}[c]{@{}c@{}}{Outcome }\\ {Variable}\end{tabular}
& \textit{Coef.} & $R^2_{Y \sim D | \mathbf{X}}$ & $RV_{q=1}$ & $RV_{q=1, \alpha\text{=0.05}}$ & $R^2_{D \sim Z_k | \mathbf{X}}$ & $R^2_{Y \sim Z_k | \mathbf{X}, D}$ \\
\hline
\begin{tabular}[c]{@{}c@{}}{Video Creation }\\ {Volume (log1p)}\end{tabular}
& 0.5548$^{***}$ & 0.0862 & 0.2635 & 0.2425 & 0.0488 & 0.0224 \\ \hline
\begin{tabular}[c]{@{}c@{}}{Total Valid} \\ {Views (log1p)}\end{tabular}
& 0.0836 & 0.0004 & 0.0198 & 0.0000 & 0.0488 & 0.0105 \\ \hline 
\begin{tabular}[c]{@{}c@{}}{Total Full}\\ {Views (log1p)}\end{tabular}
& -0.1484$^{*}$ & 0.0013 & 0.0355 & 0.0077 & 0.0488 & 0.0114 \\
\bottomrule
\end{tabular}
\label{tab:sensitivity_creation}
\end{table}

\subsubsection{Interpretation of Results}

We applied this sensitivity framework across the four primary analytical dimensions. The results consistently validate the robustness of our core findings, confirming both the substantial behavioral shifts and the structural algorithmic responses.


\begin{itemize}
    \item \textbf{Creation-side analysis (Supplementary Table~\ref{tab:sensitivity_creation}):} 
    The sensitivity analysis differentiates the robust impact on productivity from the marginal differences in aggregate returns. The estimated increase in Video Creation Volume is highly robust ($RV_{q=1} = 0.2635$); an unobserved confounder would need to exert an influence more than five times greater than our maximal observed benchmark ($R^2_{D \sim Z|\textbf{X}} = 0.0488$, $R^2_{Y \sim Z|\textbf{X},D} = 0.0224$) to eliminate this effect. Conversely, the minimal $RV$ values for Total Valid Views ($RV_{q=1} = 0.0004$) and Total Full Views ($ RV_{q=1} = 0.0013$) indicate that any estimated differences in overall engagement are highly susceptible to unobserved confounding. This lack of a robust divergence in aggregate returns naturally aligns with our baseline observation: while AIGC adoption systematically scales up production volume, it does not yield a robust advantage in total engagement returns.

\item \textbf{Consumption-side analysis (Supplementary Table~\ref{tab:sensitivity_consumption}):} The estimated negative effect of the AIGC on consumer engagement is robust to unobserved confounding. For the Valid-View Rate, while the treatment assignment explains a modest proportion of the residual variance ($R^2_{Y \sim D|\textbf{X}} = 0.0188$), the corresponding robustness value is $RV_{q=1} = 0.1292$. This required threshold is two orders of magnitude larger than the explanatory power of the strongest observed covariate ($R^2_{D \sim Z|\textbf{X}} = 0.0013$). Consistent patterns across the Full-View Rate and View Duration suggest that the observed reduction in consumer preference for AIGC is unlikely to be driven by omitted variables.

\item \textbf{Algorithmic distribution analysis (Supplementary Table~\ref{tab:sensitivity_video}):} The estimated reduction in AIGC exposure by the algorithm
is robust to potential confounding. The $RV_{q=1}$ thresholds for 31-day Cumulative Exposure and Days to Reach 90\% Exposure are $0.3224$ and $0.3344$, respectively. These values are substantially larger than the explanatory power of the primary observed pre-upload characteristics ($R^2 \approx 0.0005$). This margin indicates that the algorithmic downscaling of AIGC is well-supported, as an unobserved confounder would need a notably larger impact than the known variables to invalidate the results.

    \item \textbf{Algorithmic mechanism comparison (Supplementary Table~\ref{tab:sensitivity_personalize}):} In assessing the moderating role of algorithmic designs, the $RV_{q=1}$ for AIGC Exposure Share is estimated at $0.0948$. By comparison, the strongest empirical benchmark from consumer historical profiles yields partial $R^2$ values of $0.0210$ ($R^2_{D \sim Z|\textbf{X}}$) and $0.0050$ ($R^2_{Y \sim Z|\textbf{X},D}$). Because the theoretical threshold for an unobserved confounder exceeds the strongest observed benchmark by more than a factor of four, the suppressive effect of population feedback-driven algorithms on AIGC visibility is maintained.

\end{itemize}

\begin{table}[htbp]
\centering
\caption{\textbf{Sensitivity analysis for consumption-side engagement outcomes.} \textit{Coef.} reports the estimated regression coefficient of the treatment (the viewed video being AIGC) for each respective outcome (consumer engagement depth). $R^2_{Y \sim D | \mathbf{X}}$ denotes the proportion of residual outcome variance explained by the treatment. $RV_{q=1}$ and $RV_{q=1, \alpha=0.05}$ represent the robustness values required for an unobserved confounder to reduce the estimate to zero or render it statistically insignificant at the 5\% level, respectively. The reference columns report the maximum explanatory power of the observed covariates ($Z_k$) on the treatment ($R^2_{D \sim Z_k | \mathbf{X}}$) and the outcome ($R^2_{Y \sim Z_k | \mathbf{X}, D}$), identified by iterating through all candidates to establish an empirical benchmark.
*** $p < 0.001$, ** $p < 0.01$, * $p < 0.05$.
}
\begin{tabular}{l|cccccc}
\hline
\begin{tabular}[c]{@{}c@{}}{Outcome }\\ {Variable}\end{tabular}
& \textit{Coef.} & $R^2_{Y \sim D | \mathbf{X}}$ & $RV_{q=1}$ & $RV_{q=1, \alpha=0.05}$ & $R^2_{D \sim Z_k | \mathbf{X}}$ & $R^2_{Y \sim Z_k | \mathbf{X}, D}$ \\
\hline
\begin{tabular}[c]{@{}c@{}}{Valid-View }\\ {Rate}\end{tabular}
& -0.0767$^{***}$ & 0.0188 & 0.1292 & 0.1236 & 0.0013 & 0.0010 \\  \hline
\begin{tabular}[c]{@{}c@{}}{Full-View }\\ {Rate}\end{tabular}
& -0.0266$^{***}$ & 0.0065 & 0.0774 & 0.0715 & 0.0013 & 0.0008 \\ \hline
\begin{tabular}[c]{@{}c@{}}{View }\\ {Duration (s)}\end{tabular}
& -2.2292$^{***}$ & 0.0071 & 0.0810 & 0.0751 & 0.0013 & 0.0006 \\
\hline
\end{tabular}
\label{tab:sensitivity_consumption}
\end{table}

\begin{table}[htbp]
\centering
\caption{\textbf{Sensitivity analysis for algorithmic distribution outcomes.} \textit{Coef.} reports the regression coefficient of the treatment (the video being AIGC) for each respective outcome(algorithmic exposure metrics). $R^2_{Y \sim D | \mathbf{X}}$ denotes the proportion of residual outcome variance explained by the treatment. $RV_{q=1}$ and $RV_{q=1, \alpha=0.05}$ represent the robustness values required for an unobserved confounder to reduce the estimate to zero or render it statistically insignificant at the 5\% level, respectively. The reference columns report the maximum explanatory power of the observed covariates ($Z_k$) on the treatment ($R^2_{D \sim Z_k | \mathbf{X}}$) and the outcome ($R^2_{Y \sim Z_k | \mathbf{X}, D}$), identified by iterating through all candidates to establish an empirical benchmark.
*** $p < 0.001$, ** $p < 0.01$, * $p < 0.05$.}
\begin{tabular}{p{2.8cm}|cccccc}
\hline
Outcome Variable & \textit{Coef.} & $R^2_{Y \sim D | \mathbf{X}}$ & $RV_{q=1}$ & $RV_{q=1, \alpha=0.05}$ & $R^2_{D \sim Z_k | \mathbf{X}}$ & $R^2_{Y \sim Z_k | \mathbf{X}, D}$ \\
\hline
31-day Cumulative Exposure (log1p) & -1.7910$^{***}$ & 0.1330 & 0.3224 & 0.3202 & 0.0005 & 0.0021 \\ \hline
Days to Reach 90\% 31-day Exposure & -5.3077$^{***}$ & 0.1438 & 0.3344 & 0.3322 & 0.0005 & 0.0005 \\
\hline
\end{tabular}
\label{tab:sensitivity_video}
\end{table}

\begin{table}[htbp]
\centering
\caption{\textbf{Sensitivity analysis for algorithmic mechanism comparison.} \textit{Coef.} reports the regression coefficient of the treatment (distribution algorithm type: 1 = individual feedback-driven, 0 = population feedback-driven) for the outcome (AIGC exposure share). $R^2_{Y \sim D | \mathbf{X}}$ denotes the proportion of residual outcome variance explained by the treatment. $RV_{q=1}$ and $RV_{q=1, \alpha=0.05}$ represent the robustness values required for an unobserved confounder to reduce the estimate to zero or render it statistically insignificant at the 5\% level, respectively. The reference columns report the maximum explanatory power of the observed covariates ($Z_k$) on the treatment ($R^2_{D \sim Z_k | \mathbf{X}}$) and the outcome ($R^2_{Y \sim Z_k | \mathbf{X}, D}$), identified by iterating through all candidates to establish an empirical benchmark.
*** $p < 0.001$, ** $p < 0.01$, * $p < 0.05$.
}

\begin{tabular}{l|cccccc}
\hline
Outcome Variable & \textit{Coef.} & $R^2_{Y \sim D | \mathbf{X}}$ & $RV_{q=1}$ & $RV_{q=1, \alpha=0.05}$ & $R^2_{D \sim Z_k | \mathbf{X}}$ & $R^2_{Y \sim Z_k | \mathbf{X}, D}$ \\
\hline
\begin{tabular}[c]{@{}c@{}}{AIGC }\\ {Exposure Share}\end{tabular}
& 0.0202$^{***}$ & 0.0098 & 0.0948 & 0.0756 & 0.0210 & 0.0050 \\
\hline
\end{tabular}
\label{tab:sensitivity_personalize}
\end{table}

\subsection{Decomposition of AIGC Creators' Output}
\label{SI:DecompositionCreator}

In the main text, we observed that AIGC creators demonstrate higher overall productivity compared to HGC creators. To better understand the underlying drivers of this increased production scale, we decompose the total output of AIGC creators into their distinct AIGC and HGC video contributions.

\begin{figure}[h]
    \centering
    \includegraphics[width=0.95\linewidth]{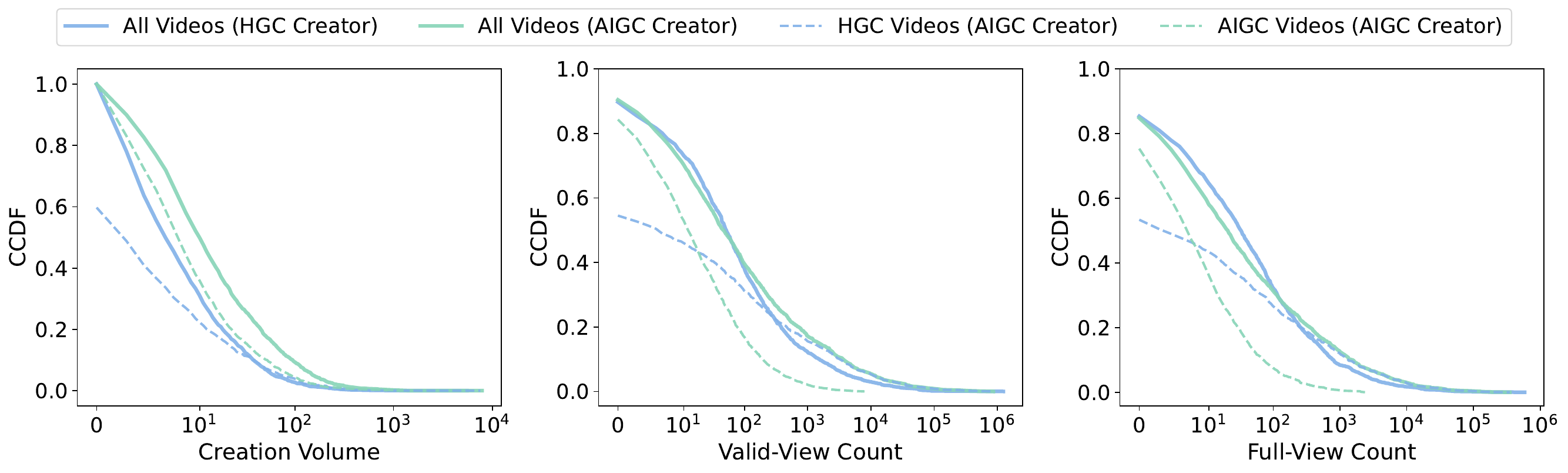}
    \caption{\textbf{Decomposition of creation volume and engagement returns for AIGC creators in creator-side comparison.}
    The complementary cumulative distribution functions (CCDFs) compare the production and engagement metrics between HGC creators (solid blue line) and AIGC creators (solid green line). To isolate the source of AIGC creators' productivity and engagement, their metrics are further decomposed into their AIGC video contributions (dashed green line) and HGC video contributions (dashed blue line). The panels illustrate (left) Creation Volume, (middle) Valid-View Count, and (right) Full-View Count.
    }
    \label{fig:si_creator}
\end{figure}

As illustrated in the left panel of Supplementary Fig.~\ref{fig:si_creator}, for the AIGC creator group, the CCDF curve for AIGC videos (dashed green line) is positioned to the right of their HGC video curve (dashed blue line). Moreover, their HGC video production curve is visibly lower than that of the baseline HGC creators (solid blue line). This confirms our assertion that the elevated creation volume among AIGC creators is primarily driven by the rapid scaling of AIGC videos, rather than an enhancement in their HGC production capacity.

Furthermore, we examine the decomposition of aggregate engagement returns—Valid-View Count (middle panel) and Full-View Count (right panel). While AIGC creators achieve overall engagement returns comparable to those of HGC creators, the decomposed curves show that their HGC videos (dashed blue line) continue to contribute a substantial portion of their total views. This suggests that within the AIGC creator category, traditional HGC remains a significant factor in maintaining engagement levels despite the increased scale of AIGC uploads.

\section{Time-series Regression Analysis}

\subsection{Granger Causality Test}
\label{SI:Granger}
To systematically investigate the dynamic temporal relationships between content supply scale and algorithmic exposure, we first employed the Granger causality test~\cite{Granger} to establish the directional precedence—specifically, whether an expansion in content supply triggers algorithmic exposure adjustments, or vice versa. To use the Granger causality test, we need to ensure the stationarity of the
time-series data, here we provide the unit root test to check it, and then we provide additional results to show the robustness of our results.

\subsubsection{Stationarity and Unit Root Tests}
To conduct the Granger causality test, it is imperative to ensure the stationarity of the time-series data to preclude the risk of spurious regression. We performed rigorous unit root testing on the first-differenced variables using both the Augmented Dickey-Fuller (ADF)~\cite{ADF} and the Kwiatkowski-Phillips-Schmidt-Shin (KPSS) tests~\cite{KPSS}. 

As detailed in Supplementary Table~\ref{tab:granger_diff}, the ADF test statistics strictly reject the null hypothesis of a unit root ($p < 0.001$) across all variables. Concurrently, the KPSS test fails to reject the null hypothesis of stationarity ($p>0.05$) for all variables. 
This cross-validation robustly confirms that our key variables—including daily AIGC/HGC supply scales and their corresponding median exposures—are stationary in their first differences, thereby satisfying the prerequisite for the subsequent Granger causality framework.

\begin{table}[ht]
\centering
\caption{\textbf{Unit root tests for variables in first differences.} 
This table presents the Augmented Dickey-Fuller (ADF) and Kwiatkowski-Phillips-Schmidt-Shin (KPSS) test results for all variables after first differencing. The test statistics are reported with p-values in parentheses. For the ADF test, the rejection of the null hypothesis indicates stationarity; for the KPSS test, the failure to reject the null hypothesis indicates stationarity. The results confirm that all variables are stationary in their first differences, satisfying the prerequisite for the subsequent Granger causality analysis. *** $p < 0.001$, ** $p < 0.01$, * $p < 0.05$.}
\begin{tabular}{l|cc}
\hline
\textbf{Variables (First Difference)} & \textbf{ADF Test Statistic} & \textbf{KPSS Test Statistic} \\ \hline
Median AIGC Exposure (31-day)         & -15.76 ($<0.001$)*** & 0.06 (0.100)                 \\
Median HGC Exposure (31-day)          & -7.06 ($<0.001$)*** & 0.04 (0.100)                 \\
Median AIGC Exposure (1-day)          & -17.23 ($<0.001$)*** & 0.10 (0.100)                 \\
Median HGC Exposure (1-day)           & -11.84 ($<0.001$)*** & 0.05 (0.100)                 \\
AIGC Supply Scale                     & -14.45 ($<0.001$)*** & 0.07 (0.100)                 \\
HGC Supply Scale                      & -6.49 ($<0.001$)*** & 0.09 (0.100)                 \\ 
\hline
\end{tabular}
\label{tab:granger_diff}
\end{table}

\begin{table}
\centering
\caption{\textbf{Granger causality test results for content supply scale and algorithmic exposure.} 
This table presents the Granger causality tests evaluating the directional relationship between content supply scale and algorithmic exposure (1-day and 31-day cumulative) for newly uploaded AIGC (Panel A) and HGC (Panel B) across multiple time lags (1 to 6 days). The results demonstrate a significant forward causality from AIGC supply scale to exposure with no significant reverse causality, whereas HGC shows no significant relationships. *** $p < 0.001$, ** $p < 0.01$, * $p < 0.05$, ns: not significant.}
\begin{tabular}{c|ccc|ccc|ccc}
\toprule
\multirow{2}{*}{\textbf{Lag}} & \multicolumn{3}{c|}{\textbf{Scale $\rightarrow$ Exposure (1d)}
}
& \multicolumn{3}{c|}{\textbf{Exposure (1d) $\rightarrow$ Scale}} & \multicolumn{3}{c}{\textbf{Scale $\rightarrow$ Exposure (31d)}} \\
\cmidrule{2-10}
 & \textbf{F-Stat} & \textbf{P-Value} & \textbf{Sig.} & \textbf{F-Stat} & \textbf{P-Value} & \textbf{Sig.} & \textbf{F-Stat} & \textbf{P-Value} & \textbf{Sig.} \\ \midrule
\multicolumn{10}{c}{\textbf{Panel A: AIGC}} \\ \midrule
1 & 0.700 & 0.404 & ns & 0.018 & 0.892 & ns & 2.787 & 0.105 & ns \\
2 & 1.423 & 0.251 & ns & 0.318 & 0.693 & ns & 3.318 & 0.052 & ns \\
3 & 6.212 & 0.012 & * & 0.339 & 0.752 & ns & 3.383 & 0.035 & * \\
4 & 5.810 & 0.009 & ** & 0.479 & 0.691 & ns & 3.322 & 0.022 & * \\
5 & 5.178 & 0.008 & ** & 0.373 & 0.811 & ns & 3.352 & 0.009 & ** \\
6 & 4.995 & 0.003 & ** & 0.667 & 0.634 & ns & 2.911 & 0.015 & * \\ \midrule
\multicolumn{10}{c}{\textbf{Panel B: HGC}} \\ \midrule
1 & 0.064 & 0.800 & ns & 0.064 & 0.800 & ns & 0.013 & 0.907 & ns \\
2 & 1.661 & 0.200 & ns & 0.072 & 0.930 & ns & 0.890 & 0.423 & ns \\
3 & 1.118 & 0.362 & ns & 0.068 & 0.973 & ns & 1.056 & 0.399 & ns \\
4 & 0.904 & 0.455 & ns & 0.067 & 0.993 & ns & 0.979 & 0.432 & ns \\
5 & 1.382 & 0.255 & ns & 0.310 & 0.896 & ns & 1.644 & 0.172 & ns \\
6 & 1.213 & 0.321 & ns & 0.468 & 0.827 & ns & 1.639 & 0.176 & ns \\ \bottomrule
\end{tabular}
\label{tab:granger}
\end{table}

\subsubsection{Statistic Significance and Reverse Checking for Granger Causality Test}

Having satisfied the stationarity prerequisite, we conducted the Granger Causality Test to evaluate whether an expansion in supply triggers algorithmic exposure adjustments. 




Methodologically, our Granger causality analysis evaluates the predictive power of lagged supply on future exposure via a joint F-test. This test compares the unrestricted full model against a restricted model ($H_{0}: \gamma_{1} = \dots = \gamma_{p} = 0$), where a significant F-statistic indicates that incorporating historical supply significantly improves exposure predictions. 
Supplementary Table~\ref{tab:granger} presents the F-statistics across various lag specifications ($p$) for both 1-day and 31-day cumulative exposures.

For AIGC (Supplementary Table~\ref{tab:granger}, Panel A), the tests demonstrate a highly significant forward causality. While low-order lags (lags 1 and 2) are non-significant—reflecting a brief algorithmic buffer period required to accumulate and react to supply shocks—the relationship becomes robustly significant at higher-order lags (e.g., $p=0.012$ for 1-day exposure at lag 3).
The reverse causality yields no statistical significance, confirming a strictly unidirectional relationship driven by supply.
Furthermore, this forward causality extends significantly to the 31-day cumulative exposure, indicating the algorithm's moderation spans the content's extended lifecycle. 
Reverse causality for the 31-day metric is omitted, as future accumulation lacks the temporal precedence required for Granger lead significance.

Besides, we conducted a parallel test for HGC. In this specification, we set the target variable as the median exposure of HGC ($\Delta E_t^{HGC}$) and the focal explanatory variable as the HGC supply scale ($\Delta S_t^{HGC}$), while explicitly controlling for the AIGC supply scale ($\Delta S_t^{AIGC}$).
As shown in Supplementary Table \ref{tab:granger} (Panel B), the HGC supply scale exhibits no significant directional relationship with its corresponding exposure across all lags and in either direction. 
These results imply that AIGC exposure is significantly more responsive to changes in its supply scale compared to HGC. 
This pronounced responsiveness likely stems from AIGC’s heavier reliance on the platform’s initial algorithmic traffic allocation to gain visibility, making the system’s exposure allocation highly responsive to its rapid supply expansion compared to HGC.



\subsection{Dynamic OLS (DOLS) Regression Estimates}

Having established through the Granger causality test that the expansion of AIGC supply precedes adjustments in algorithmic exposure, we have further employed Dynamic Ordinary Least Squares (DOLS)~\cite{stock1993simple} models to examine the more detailed associations between the content supply scale and algorithmic exposure distribution as discussed. Here, we provide additional details, including the regression equation and more comprehensive results with statistical significance testing.

\begin{table}[]
\centering
\caption{
\textbf{Dynamic OLS (DOLS) regression estimates: associations between AIGC supply scale and exposure tier distribution.}
This table reports the associations between the AIGC supply scale and the proportion of newly uploaded AIGC videos entering specific algorithmic exposure tiers. For readability, the estimated coefficients and their 95\% bias-corrected and accelerated (BCa) confidence intervals (in brackets) for the AIGC supply scale are multiplied by $10^6$. The HGC supply scale is included to control for platform-level volume dynamics; its coefficients are omitted for brevity. Optimal DOLS lag orders ($p$) were selected via the Akaike Information Criterion (AIC)~\cite{akaike2003new}. The Shin test evaluates residual stationarity; the high p-values (failure to reject the null) across all models confirm stationarity, precluding spurious regression. *** $p < 0.001$, ** $p < 0.01$, * $p < 0.05$.
}
\begin{tabular}{ccccc}
\toprule
\textbf{Exposure Tier} & \textbf{Low} & \textbf{ Med.-Low} & \textbf{ Med.-High} & \textbf{ High} \\
\textbf{Variables} & \textbf{($\le$10 views)} & \textbf{(11-100 views)} & \textbf{(101-1K views)} & \textbf{($>$1K views)} \\ 
\midrule


\begin{tabular}[c]{@{}c@{}}{AIGC Supply }\\ {Scale ($\times 10^6$)}\end{tabular}

& 1.43*** & -0.02 & -1.37*** & -0.08*** \\
                                  & [1.10, 1.76] & [-0.25, 0.20] & [-1.63, -1.12] & [-0.09, -0.06] \\ \midrule
HGC Scale Control                 & Yes & Yes & Yes & Yes \\ 
DOLS Lag Order              & 0 & 0 & 1 & 0 \\
Shin Test Statistic               & 1.10 & 1.08 & 0.33 & 0.85 \\
Shin Test P-value                 & 0.990 & 0.990 & 0.982 & 0.990 \\
Adjusted R-squared                & 0.448 & 0.038 & 0.609 & 0.509 \\ \bottomrule
\end{tabular}
\label{tab:dols_tiered_regression}
\end{table}

\subsubsection{Association between AIGC Supply Scale and Exposure Tier Distribution}





In our analysis of the initial visibility allocation of newly uploaded AIGC content across different exposure tiers, we instantiate the DOLS model as:

$$\pi_{t}^{k} = \alpha^{k} + \beta^{k} S_{t}^{AIGC} + \theta^{k} S_{t}^{HGC} + \sum_{j=-p}^{p} \gamma_{j}^{k} \Delta S_{t-j}^{AIGC} + \epsilon_{t}^{k} + \delta^{k} D_{t},$$
where $\pi_{t}^{k}$ is the proportion of new AIGC videos entering exposure tier $k$ (e.g., $\le10$ views, $11\text{--}100$ views, etc.), $S_{t}^{AIGC}$ is the absolute supply scale of AIGC, and the HGC supply scale ($S_{t}^{HGC}$) is included as a control variable to account for volume dynamics. 
To ensure the validity of our model, we utilize the Shin test~\cite{shin1994residual} to examine residual stationarity, a critical step that prevents the risk of spurious regression.
Furthermore, given the time-series nature of our data, the confidence intervals (CIs) and $p$-values are generated using block bootstrapping~\cite{hall1995blocking} to robustly account for potential serial correlation and heteroskedasticity.  
Specifically, the bootstrap block size is set to 7, serving as an approximation of $T^{1/3}$ for our one-year observation period ($T=365$ days), which is the recommended optimal block length $T^{1/3}$ \cite{hall1995blocking}. Notably, for the following other DOLS experiments, the Shin test and block bootstrapping–based analyses are performed by default, and the corresponding results are reported.

Supplementary Table \ref{tab:dols_tiered_regression} reports the DOLS estimates for the proportion of newly uploaded AIGC videos entering four specific algorithmic exposure tiers. 
The results indicate that an increase in the AIGC supply scale is significantly and positively associated with a higher proportion of new AIGC videos falling into the lowest exposure tier ($\le10$ views). 
Conversely, an expanded AIGC supply demonstrates significant negative associations with entry into the medium-high tier ($101\text{--}1000$ views) and the high exposure tier ($>1000$ views), while showing no significant relationship with the medium-low tier ($11\text{--}100$ views). 
These estimates objectively demonstrate that the algorithmic distribution mechanism systematically downscales the initial exposure allocated to newly uploaded AIGC items as its supply expands.

\subsubsection{Algorithmic Exposure Response to Scale-over-Preference Dynamics}

Having established through matching analyses that the algorithmic distribution mechanism aligns with consumer preference by assigning lower exposure to AIGC, and through time-series regressions that it systematically downscales initial visibility as AIGC supply expands, we then turned to how the system navigates the core ecological tension: the SoP dynamic.
Specifically, we employed the DOLS model to examine the associations of relative supply scale ($S$) and relative consumer preference ($P$) with the relative median exposure of AIGC. 
Building upon the general DOLS framework, the specific model is instantiated as follows:

$$E_{t} = \alpha + \beta_{S} S_{t} + \beta_{P} P_{t} + \sum_{j=-p}^{p} \gamma_{j,S} \Delta S_{t-j} + \sum_{j=-p}^{p} \gamma_{j,P} \Delta P_{t-j} + \epsilon_{t} + \delta D_{t},$$
where $E_{t}$ denotes the median exposure of AIGC relative to HGC, $S_{t}$ represents the relative supply scale of AIGC to HGC, and $P_{t}$ reflects the relative consumer preference for AIGC based on valid-view ratios.

\begin{table}[t]
\centering
\caption{
\textbf{Dynamic OLS (DOLS) regression estimates: algorithmic exposure response to scale-over-preference dynamics.} 
This table reports the associations of relative consumer preference and relative supply scale with the relative algorithmic exposure of AIGC to HGC. Consistent with Fig. 2(c) in the main text, the results demonstrate a dual moderation mechanism: holding supply scale constant, algorithmic exposure increases significantly with stronger consumer preference; conversely, holding preference constant, exposure decreases significantly as supply scale expands. Estimated coefficients are presented with their 95\% bias-corrected and accelerated (BCa) confidence intervals in brackets. 
The DOLS lag order was optimally selected via the Akaike Information Criterion (AIC). 
The Shin test evaluates residual stationarity; the high p-value confirms stationarity, precluding spurious regression. *** $p < 0.001$, ** $p < 0.01$, * $p < 0.05$.}
\begin{tabular}{lc}
\hline
\textbf{Variables} & 
\begin{tabular}[c]{@{}c@{}}{ \textbf{Relative Median Exposure} }\\ {\textbf{ (AIGC to HGC)}}\end{tabular}

\\ \hline
Relative Consumer Preference for AIGC & 2.683*** \\
                                      & [1.663, 3.725] \\
Relative Supply Scale of AIGC         & -0.936*** \\
                                      & [-1.090, -0.794] \\ \hline
DOLS Lag Order (AIC)                  & 4 \\
Shin Test Statistic                   & 0.138 \\
Shin Test P-value                     & 0.846 \\
Adjusted R-squared                    & 0.639 \\ \hline
\end{tabular}

\label{tab:dols_sop}
\end{table}

Supplementary Table \ref{tab:dols_sop} reports the regression estimates. The results reveal a significant dual-moderation mechanism: holding the supply scale constant, relative consumer preference is positively associated with relative exposure ($\beta_{P} = 2.683$, $p < 0.001$). Conversely, holding preference constant, an increase in relative supply scale is significantly and negatively associated with relative exposure ($\beta_{S} = -0.936$, $p < 0.001$). 
The Shin test for this model yields a $p$-value of 0.846, failing to reject the null hypothesis and confirming residual stationarity. These findings provide empirical evidence that the algorithm moderates the SoP tension by systematically adjusting visibility based on the relative performance of supply and preference signals.

\subsubsection{Elasticities of Supply Scale on Ecological Outcomes}

To evaluate the downstream consequences of algorithmic moderation on the content ecosystem, we further estimate the elasticities of creator and consumer outcomes with respect to the supply scale. 
Given the distinct mechanisms affecting creators and consumers, we specify two separate log-log DOLS models. 
In a log-log specification, the estimated coefficients represent elasticities—that is, the percentage change in the outcome variable associated with a 1\% increase in supply.

\begin{table}[t]
\centering
\caption{\textbf{Dynamic OLS (DOLS) regression estimates: elasticities of supply scale on creator engagement returns. }
This table reports the elasticities derived from a log-log DOLS specification, corresponding to Fig. 2(d) in the main text. The dependent variables are the aggregate full-view and valid-view counts for AIGC and HGC creators, respectively. 
The coefficient of interest is the "Own-type Supply Scale," representing the elasticity of engagement with respect to the supply of the same content type (e.g., AIGC engagement relative to AIGC supply).
The cross-type supply scale (e.g., HGC supply in AIGC models) is included as a control in all specifications.
Estimated elasticities are presented with 95\% bias-corrected and accelerated (BCa) confidence intervals in brackets. 
The DOLS lag orders were selected via the Akaike Information Criterion (AIC). 
The high Shin test p-values confirm residual stationarity across all models. *** $p < 0.001$, ** $p < 0.01$, * $p < 0.05$.}
\begin{tabular}{lcccc}
\hline
 \textbf{Outcome Types} & \multicolumn{2}{c}{\textbf{AIGC Creator Outcomes}} & \multicolumn{2}{c}{\textbf{HGC Creator Outcomes}} \\  \midrule
 & \textbf{ Full-View} & \textbf{ Valid-View} & \textbf{ Full-View} & \textbf{Valid-View} \\ 
\textbf{Variables (Log)} & \textbf{Count} & \textbf{Count} & \textbf{Count} & \textbf{Count} \\ \midrule
Own-type Scale Elasticity & 0.191 & 0.132 & 0.716*** & 0.508*** \\
            & [-0.039, 0.364] & [-0.046, 0.268] & [0.325, 1.092] & [0.211, 0.802] \\ \midrule
Cross-type Scale Control & Yes & Yes & Yes & Yes \\ 
DOLS Lag Order (AIC)     & 4 & 4 & 0 & 0 \\
Shin Test Statistic      & 1.22 & 0.36 & 1.80 & 1.35 \\
Shin Test P-value        & 0.990 & 0.987 & 0.990 & 0.990 \\
Adjusted R-squared       & 0.104 & 0.114 & 0.118 & 0.117 \\ \bottomrule
\end{tabular}
\label{tab:dols_creator}
\end{table}

\vspace{+10pt}
\noindent\textbf{Creator Outcomes: Elasticities of Supply Scale on Engagement Returns}

For the creator side, we examine how the supply scale of a specific content category (e.g., AIGC) impacts its own-type overall engagement returns, controlling for the supply scale of the opposing category (e.g., HGC). We measure creator outcomes as the category-specific overall engagement return, which is the sum of full-view and valid-view counts received by all newly uploaded videos within that category. The model is instantiated as:
\begin{align}
\ln(ER_t^{AIGC}) &= \alpha + \beta \ln(S_t^{AIGC}) + \theta \ln(S_t^{HGC}) + \sum_{j=-p}^p \gamma_j \Delta \ln(S_{t-j}^{AIGC}) + \epsilon_t + \delta D_t \\
\ln(ER_t^{HGC}) &= \alpha' + \beta' \ln(S_t^{HGC}) + \theta' \ln(S_t^{AIGC}) + \sum_{j=-p}^p \gamma'_j \Delta \ln(S_{t-j}^{HGC}) + \epsilon'_t + \delta' D_t
\end{align}
where $ER_t^{AIGC}$ and $ER_t^{HGC}$ represent the category-specific overall engagement return (measured by aggregate full-view and valid-view counts of newly uploaded videos) for AIGC creators and HGC creators, respectively, and $S_t^{AIGC}$ (or $S_t^{HGC}$) is the absolute supply scale of AIGC (or HGC). The coefficient $\beta$ (or $\beta'$) captures the own-type elasticity.

Supplementary Table \ref{tab:dols_creator} reports these estimates. The results indicate that the elasticity of AIGC engagement returns with respect to its own supply scale is significantly positive but small ($\beta = 0.132$ for full-view; $\beta = 0.191$ for valid-view, $p < 0.05$). In contrast, HGC creators exhibit substantially larger elasticities ($\beta^\prime = 0.716$ for full-view; and $\beta^\prime = 0.508$ for valid-view; $p < 0.001$). 
This dampened elasticity aligns with the algorithmic downscaling mechanism identified earlier.
It implies that the algorithm's restriction on AIGC visibility effectively bounds the aggregate gains of creators: when scaling content production proportionally, AIGC creators capture substantially smaller relative returns in engagement compared to their HGC counterparts.

\begin{table}[t]
\centering
\caption{\textbf{Dynamic OLS (DOLS) regression estimates: elasticities of supply scale on consumer engagement depth.} 
This table reports the elasticities from a log-log DOLS specification. Dependent variables are the overall full-view and valid-view rates. Both AIGC and HGC supply scales are included to examine their respective associations.
Consistent with Fig. 2e in the main text, elasticities for both supply scales are near zero and statistically insignificant, indicating that algorithmic distribution stabilizes user engagement despite massive AIGC expansion.
Estimated elasticities are presented with 95\% BCa confidence intervals in brackets (constant omitted). Optimal DOLS lag orders were selected via AIC. The Shin test evaluates residual stationarity; high p-values confirm stationarity, precluding spurious regression.
*** $p < 0.001$, ** $p < 0.01$, * $p < 0.05$.
}
\begin{tabular}{lcc}
\toprule
 & \textbf{(1) Overall Full-View} & \textbf{(2) Overall Valid-View} \\
 
\textbf{Variables (Log)} & \textbf{Rate} & \textbf{Rate} \\ \midrule
AIGC Supply  & 0.026 & 0.025 \\
Scale (Elasticity)      & [-0.054, 0.117] & [-0.028, 0.086] \\ \midrule
HGC Supply  & 0.256 & 0.040 \\
Scale (Elasticity)      & [-0.114, 0.632] & [-0.198, 0.301] \\ \midrule
DOLS Lag Order (AIC) & 0 & 0 \\
Shin Test Statistic  & 1.80 & 1.61 \\
Shin Test P-value    & 0.990 & 0.990 \\
Adjusted R-squared   & 0.025 & 0.021 \\ \bottomrule
\end{tabular}
\label{tab:dols_elasticity_consumer}
\end{table}

\vspace{+10pt}
\noindent\textbf{Consumer Outcomes: Elasticities of Supply Scale on Engagement Depth}

For the consumer side, we examine how the expansion of both AIGC and HGC supply scales simultaneously impacts the overall consumer experience. 
The model is instantiated as:

\begin{equation*}
\begin{aligned}
\ln(ED_{t}) =\ & \alpha + \beta^{AIGC}\ln(S_{t}^{AIGC}) + \beta^{HGC}\ln(S_{t}^{HGC}) \\
& + \sum_{j=-p}^{p}\gamma_{j}^{AIGC} \Delta\ln(S_{t-j}^{AIGC}) + \sum_{j=-p}^{p}\gamma_{j}^{HGC} \Delta\ln(S_{t-j}^{HGC}) + \epsilon_{t} + \delta D_{t}
\end{aligned}
\end{equation*}
where $ED_{t}$ represents the overall consumer engagement depth (measured by the average full-view and valid-view rates across all newly uploaded videos, regardless of content type).

Supplementary Table \ref{tab:dols_elasticity_consumer} reports these estimates. The results show that the elasticities of consumer engagement depth with respect to the AIGC supply scale are near zero and statistically insignificant ($\beta = 0.026$ and $\beta = 0.025$). This indicates that despite the massive expansion of AIGC, overall consumer engagement does not degrade. This finding is consistent with the algorithmic downscaling mechanism identified earlier: by restricting the visibility of AIGC, the distribution system effectively shields the overall user experience from the influx of lower-preference content.




\end{document}